\documentclass[10pt,twocolumn,letterpaper]{article}

\usepackage[pagenumbers]{cvpr} 

\usepackage[dvipsnames]{xcolor}

\usepackage{bbding}

\usepackage{tikz}
\usetikzlibrary{patterns}
\newcommand{\myrectangle}{
    \begin{tikzpicture}
        \definecolor{mycolor}{RGB}{250,112,112}
        \fill[red, pattern=north east lines, pattern color=mycolor] (0,0) rectangle (0.4,0.2);
    \end{tikzpicture}
}

\newcommand\blfootnote[1]{%
  \begingroup
  \renewcommand\thefootnote{}\footnote{#1}%
  \addtocounter{footnote}{-1}%
  \endgroup
}

\definecolor{cvprblue}{rgb}{0.21,0.49,0.74}
\usepackage[pagebackref,breaklinks,colorlinks,citecolor=cvprblue]{hyperref}
\usepackage[accsupp]{axessibility}
\usepackage{amssymb}
\usepackage{xcolor}
\usepackage{colortbl} 

\newcommand{\fakeparagraph}[1]{\vspace{3mm}\noindent\textbf{#1}}

\usepackage[capitalize]{cleveref}
\crefname{section}{Sec.}{Secs.}
\Crefname{section}{Section}{Sections}
\Crefname{table}{Table}{Tables}
\crefname{table}{Tab.}{Tabs.}

\newcommand{\HalfCheckmark}{%
  \Checkmark\kern-1.2ex\raisebox{1ex}{\rotatebox[origin=c]{125}{\textbf{--}}}%
}

\def\paperID{} 
\def\confName{CVPR}
\def\confYear{2024}

\title{Synthesize, Diagnose, and Optimize: Towards Fine-Grained \\ Vision-Language Understanding}

\author{
Wujian~Peng$^{1,2}$~\hspace{4pt}
Sicheng~Xie$^{1,2}$~\hspace{4pt}
Zuyao~You$^{1,2}$~\hspace{4pt}
Shiyi~Lan$^{3}$~\hspace{4pt}
Zuxuan~Wu$^{1,2\dagger}$\vspace{0.1in}\\
$^{1}$Shanghai Key Lab of Intell. Info. Processing, School of CS, Fudan University \\
$^{2}$Shanghai Collaborative Innovation Center of Intelligent Visual Computing \\
$^{3}$NVIDIA
}

\begin{document}
\maketitle
\begin{abstract}
\blfootnote{$^{\dagger}$ Corresponding author.}
Vision language models (VLM) have demonstrated remarkable performance across various downstream tasks. However, understanding fine-grained visual-linguistic concepts, such as attributes and inter-object relationships, remains a significant challenge. While several benchmarks aim to evaluate VLMs in finer granularity, their primary focus remains on the linguistic aspect, neglecting the visual dimension. Here, we highlight the importance of evaluating VLMs from both a textual and visual perspective. We introduce a progressive pipeline to synthesize images that vary in a specific attribute while ensuring consistency in all other aspects. Utilizing this data engine, we carefully design a benchmark, SPEC, to diagnose the comprehension of object size, position, existence, and count. Subsequently, we conduct a thorough evaluation of four leading VLMs on SPEC. Surprisingly, their performance is close to random guess, revealing significant limitations. With this in mind, we propose a simple yet effective approach to optimize VLMs in fine-grained understanding, achieving significant improvements on SPEC without compromising the zero-shot performance. Results on two additional fine-grained benchmarks also show consistent improvements,  further validating the transferability of our approach. Code and data are available at \url{https://github.com/wjpoom/SPEC}.
\end{abstract}
\section{Introduction}
\label{sec:intro}
\begin{figure}
    \centering
    \includegraphics[width=\linewidth]{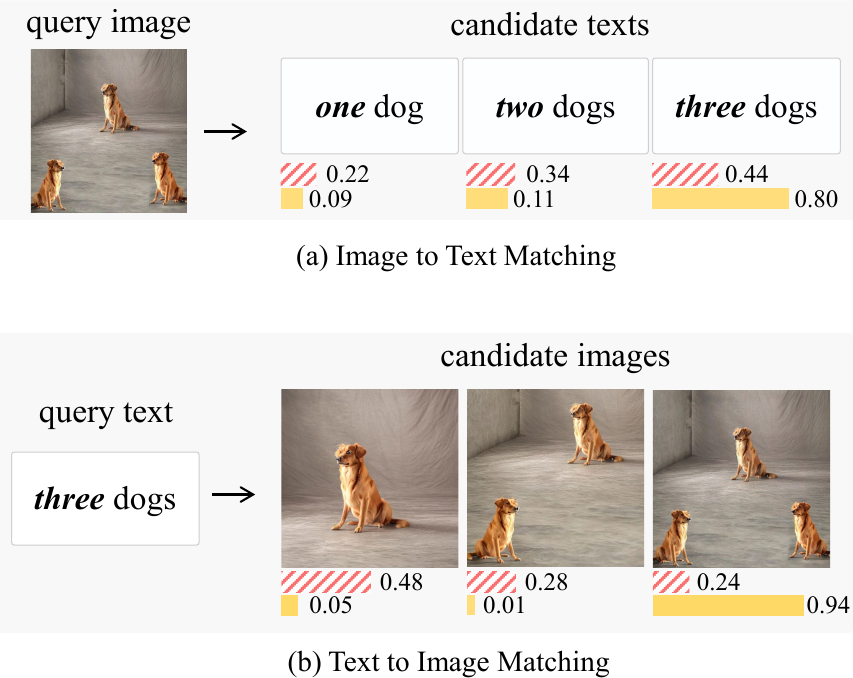}
    \caption{
    \textbf{We conduct a symmetrical assessment of VLMs in fine-grained comprehension}, considering both linguistic and visual perspectives.
    The bars in \myrectangle and \textcolor[RGB]{255, 217, 102}{\rule{0.4cm}{0.2cm}} represent the image-text matching scores for CLIP~\cite{clip} and our method, respectively. It is evident that CLIP struggles with tasks related to quantity comprehension, whereas our method significantly enhances the model in understanding fine-grained details.
    } 
    \vspace{-5mm}
    \label{fig:teaser}
\end{figure}

Vision and Language Foundation Models~(VLMs) pretrained on large-scale image-text data~\cite{clip, jia2021scaling, singh2022flava, x-vlm, li2022blip} have consistently demonstrated impressive performance across a wide range of well-established evaluating tasks, \ie image classification~\cite{deng2009imagenet}, image captioning~\cite{coco, nocaps}, visual question answering~\cite{Chen2022PaLIAJ} and cross-modal image-text retrieval~\cite{coco,flicker}. Their remarkable performance is gradually convincing the community that these currently available VLMs are almost robust and powerful enough to be transferred to a broad spectrum of downstream tasks, either through finetuning or even in a zero-shot manner.

However, recent research has shattered this captivating illusion, revealing that even state-of-the-art VLMs~\cite{clip, li2022blip, singh2022flava, x-vlm} exhibit significant limitations in understanding visual-linguistic concepts that require fine-grained compositional reasoning, especially in tasks involving object attributes or inter-object relationships~\cite{thrush2022winoground, bow_aro, vlchecklist, VALSE, Eqben}. This raises a crucial question: \textit{to what extent and in what aspects are VLMs excelling or struggling?} To answer this, previous effort evaluates fine-grained capabilities of VLMs through the image-to-text matching task, as shown in~\cref{fig:teaser}(a). This involves providing a query image and retrieving the matching text from a set of confusing candidates, differing subtly in texts. For example, when assessing the counting ability, it is crucial to ensure that quantity is the unique variable and other clues are kept the same. Therefore, a straightforward way to do so is modifying quantity and adjective words used in the texts. While manipulating texts to construct confusing candidate sets has been well-studied due to the sparsity of the text space and advancements in Large Language Models (LLMs)~\cite{bow_aro, vlchecklist, vsr, VALSE}, the visual side remains relatively under explored, largely due to the complexity of visual signals and the absence of powerful tools. We posit that exploring the visual dimension in a fine-grained manner is also essential for a comprehensive understanding of VLMs.

\begin{figure*}[ht]
    \centering
    \includegraphics[width=\linewidth]{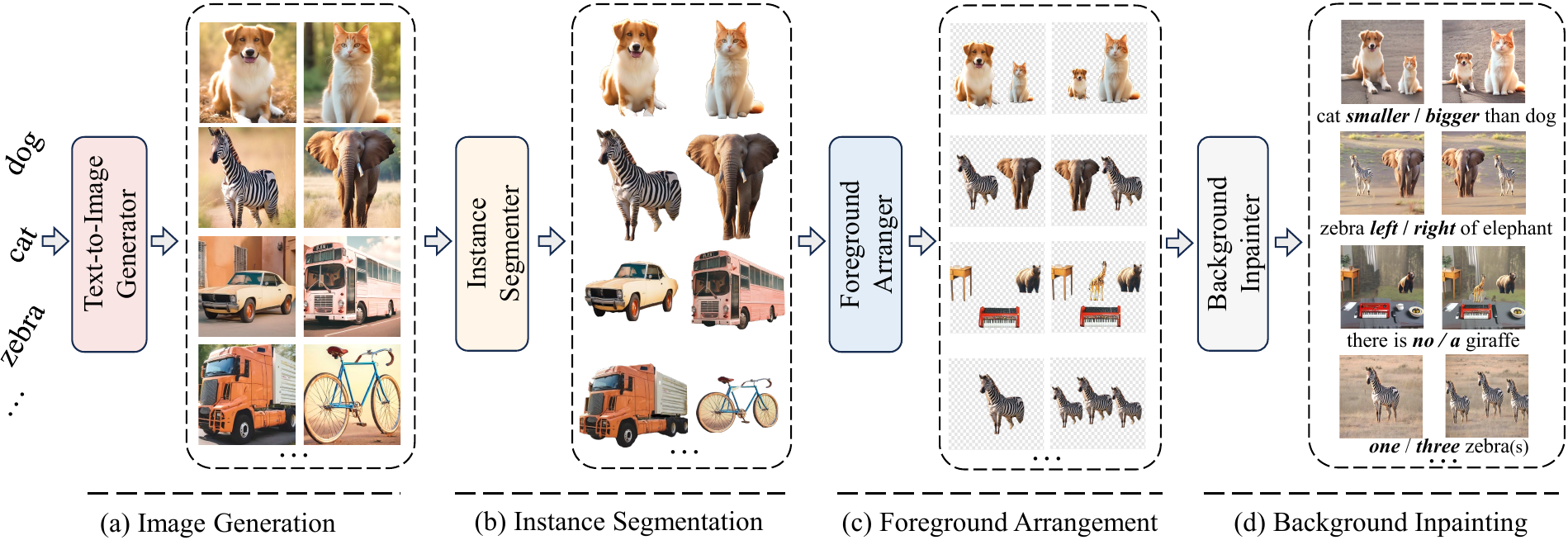}
    \caption{\textbf{The overall illustration of our data progressive construction pipeline.} We initiate the process by generating a batch of images containing a single object. Subsequently, we extract the object from the background in the images. Following that, we arrange the background-free images on a blank canvas according to specifications (with control over attributes). Finally, we meticulously fill in the missing background, ensuring consistency across candidates.}
    \vspace{-4mm}
    \label{fig:data_pipeline}
\end{figure*}

Motivated by the great progress achieved in generative models, we present an effective framework for generating high-quality image candidates that are suitable for evaluating the performance of VLMs.
This framework ensures that images within the same candidate set only differ in the specified property of interest, while all other properties remain consistent. We break down this task into several simple and manageable steps. As illustrated in~\cref{fig:data_pipeline}, we start by utilizing a text-to-image model~\cite{stable_diffusion, sdxl} to generate images featuring a single object. Then, a segmenter~\cite{sam,sam_hq} is employed to separate the objects from their backgrounds, yielding a library of foreground instance spanning various categories. From there, we select instances and arrange them on a blank canvas (manipulating attributes such as size, position, existence, and quantity of a specific object at this stage is straightforward). Finally, we use an inpainting model~\cite{sdxl, stable_diffusion} to fill the missing background portions, producing an photo-realistic image. It is worth noting that, during the inpainting process, we design a progressive background filling strategy, effectively ensuring consistency in the background across all images in the same candidate set.

Empowered by this data construction pipeline, we carefully develop a new benchmark, named as \textbf{SPEC}, to evaluate the proficiency of VLMs in comprehending fine-grained concepts including \textbf{S}ize, \textbf{P}osition, \textbf{E}xistence and \textbf{C}ounting. We systematical test four VLMs~\cite{clip, li2022blip, singh2022flava, x-vlm} on this newly created test bed. Surprisingly, even state-of-the-art models perform at chance-level, exposing significant performance deficiencies. Following this, we implement a straightforward approach to remedy this by incorporating hard negative examples~(\ie, confusing images or texts within the same candidate set) into the same training batch. This encourages the model to discern subtle differences among candidate examples, leading to a significant improvement in performance on SEPC while preserving the original zero-shot capability. Furthermore, to demonstrate the model's generalization ability, we conduct additional tests on two existing datasets~\cite{bow_aro, Eqben}, which also focus on compositional reasoning. The consistent improvement further validates that our method effectively guide the model to acquire essential and transferable comprehending abilities at a finer granularity.
Our main contributions are:
\begin{enumerate}[]
    \item \textbf{A progressive data constructing pipeline}.
    We present a progressive data construction pipeline designed for creating a candidate image set. Within each candidate set, images vary exclusively in a specified attribute while ensuring consistency across other aspects. Such data are valuable for conducting text-to-image matching tasks, as depicted in~\cref{fig:teaser}(b), offering a visual perspective for evaluating VLMs.
    \item \textbf{A carefully curated benchmark: SPEC}. 
    We meticulously craft a novel benchmark, SPEC, with a specific focus on evaluating VLMs' understanding of fine-grained visual-linguistic concepts, encompassing object size, position, existence, and count. The introduction of SPEC enables a symmetrical evaluation of VLMs from both image and text perspectives, addressing the previous lack of image-centric testing data.
    \item \textbf{A simple and effective remedy}. 
    We evaluate four VLMs on SPEC, revealing significant limitations. In response, we propose a method to enhance the understanding of fine-grained visual-linguistic concepts. Experimental results indicate notable improvement not only on SPEC but also consistent results on two additional datasets, while preserving zero-shot capability.
\end{enumerate}

\section{Related Work}
\label{sec:related_works}
\fakeparagraph{Vision and Language Models~(VLMs).} Models such as CLIP~\cite{clip}, ALIGN~\cite{jia2021scaling}, CyCLIP~\cite{cycleCLIP} CoCa~\cite{CoCa}, OmniVL~\cite{omnivl} and Open-VCLIP~\cite{weng2023transforming, wu2024building} have demonstrated impressive performance across a wide range of downstream tasks. These models include two separate unimodal encoders, each designed to extract representations for visual and textual input. To achieve alignment between the two modalities, they typically employ a huge number of visual-textual pairs for contrastive learning. By pretraining on 400M noisy data, CLIP~\cite{clip} achieves a top-1 accuracy on ImageNet-1K~\cite{deng2009imagenet} comparable to that of ResNet-50~\cite{resnet}, even though it is never specifically trained on ImageNet and is evaluated in a zero-shot manner. However, as highlighted in recent work~\cite{thrush2022winoground, bow_aro}, these advancements are primarily attributed to the simplicity of evaluation tasks which requires no reasoning or compositional capabilities. The performance of these models are limited on tasks that require fine-grained understanding~\cite{bow_aro, VALSE}.

\vspace{2mm}
\noindent\textbf{Benchmarking VLMs in Finer Granularity.} To assess the model's understanding of nuanced visual-linguistic concepts, several new benchmarks have been proposed. Winoground~\cite{thrush2022winoground} is curated by experts with a focus on compositional understanding. VALSE~\cite{VALSE} and VL-Checklist~\cite{vlchecklist} investigate several linguistic phenomena by transforming real captions into confusing alternatives. ARO~\cite{bow_aro} diagnoses VLMs in attribution, relation and ordering. Eqben~\cite{Eqben} assesses whether the model is sensitive to visual semantic changes. Most existing benchmarks solely focus on subtle textual changes~\cite{VALSE, vlchecklist, bow_aro, vsr, verb}, as crafting confusing text candidates is straightforward that can be achieved through either LLMs or simple rules. Winoground~\cite{bow_aro} and Eqben~\cite{Eqben} are most relevant to us, since we focus on minimal semantic changes in both images and texts, enabling a more comprehensive evaluation across modalities. However, the scale of Winoground is restricted by its costly curation, and the image diversity of Eqben is limited by virtual engines. In contrast, our data construction pipeline is scalable and can produce diverse images.

\vspace{2mm}
\noindent\textbf{Enhancing VLMs for Fine-grained Understanding.} To mitigate challenges in fine-grained recognition, various  approaches have been explored. Syn-CLIP~\cite{beyoundNouns} utilize data synthesized by 3D simulation engines to enhance the model's understanding of concepts beyond nouns. EQSIM~\cite{Eqben} incorporates an additional regularization loss to generalize VLMs to nuanced multimodal compositions. TSVLC~ ~\cite{teachingstruct} and ViLEM~\cite{ViLEM} introduce negative texts generated by LLMs~\cite{scao2022bloom, BERT} to inject fine-grained knowledge. Construct-VL~\cite{constructVL} addresses these challenges from a continual learning perspective. These methods compel the model to focus on subtle differences by introducing confusing texts as hard negatives. However, we argue that the absence of visual hard negatives limits its performance. Thus, we introduce hard negatives for both modalities, simultaneously enhancing the visual and textual encoders.
\section{Synthesize: Data Construction Pipeline}
\label{sec:method}
Our goal is to build a set of perplexing image, wherein each image differs solely in a specified attribute while ensuring consistency in all other aspects.  We first emphasize the importance of preserving consistency among candidate images for effective evaluation~(\cref{sec:consistency_matter}). To address this, we break down this problem and introduce a progressive data construction pipeline~(\cref{sec:data_pipeline}). Then, we carefully devise a benchmark that centers on evaluating VLMs' grasp of fine-grained visual-linguistic concepts~(\cref{sec:spec_benchmark}).

\begin{figure}
    \centering
    \includegraphics[width=0.7\linewidth]{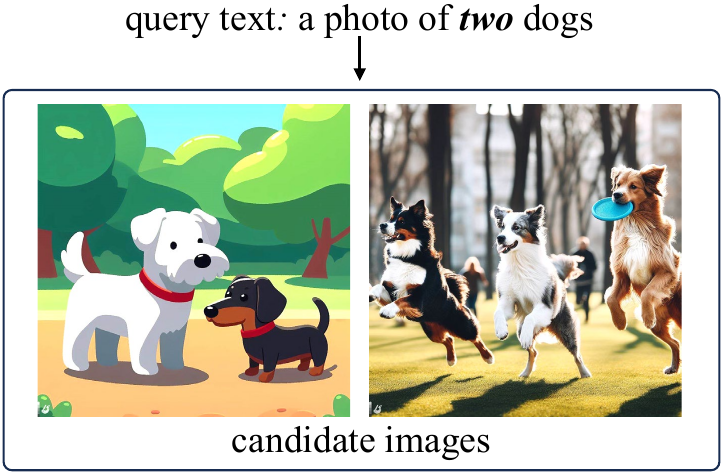}
    \caption{\textbf{Ensuring consistency among candidates is crucial to avoid ambiguity.} The images above not only differ in quantity but also show a significant variation in the appearance of the objects. Consequently, attributing the model's correctness or errors solely to the understanding of quantity is not convincing.}
    \vspace{-3mm}
    \label{fig:consistency_taser}
\end{figure}
\subsection{Importance of Candidate Consistency}
\label{sec:consistency_matter}
As illustrated in~\cref{fig:consistency_taser}, when conducting a matching task with the textual query ``a photo of two dogs'', the model might mistakenly choose the image on the right~(which actually contains three dogs). However, attributing this error solely to counting difficulty is not convincing. The model might select the right-side image due to its more photo-realistic appearance or better alignment with the word ``dog'', as the left-side image has a cartoon style. Conversely, if the model correctly selects the right-side image for the query ``a photo of three dogs'', we also cannot assert that the model is proficient in counting. These obscure ambiguities arise because there is no guarantee of the uniqueness of changing factors among candidate images during evaluation. Therefore, ensuring consistency among candidates in all aspects except the one under investigation is crucial. To this end, we propose a progressive data construction method, which will be elaborated in detail as follows.

\subsection{Progressive Data Construction}
\label{sec:data_pipeline}
The data construction framework is illustrated in~\cref{fig:data_pipeline}, comprising four progressive steps. Initially, we generate images featuring a single prominent object. Then, we isolate the foreground from the background, resulting in a library of foreground objects spanning various categories. Subsequently, we select objects from this library and arrange them on a blank canvas, adjusting their attributes and relationships. Lastly, we carefully fill in the missing background, ensuring consistency among different candidates.

\subsubsection{Generating Images with Single Objects}
We initiate the process by utilizing a generation model to obtain a collection of images, each featuring a single and prominent object corresponding to a specific category. Due to the progress in visual generation models~\cite{stable_diffusion, google_imagin, xing2023simda, vdmsurvey}, these images display a high level of photo-realistic and diverse content. In practice, we use Stable-Diffusion-XL 1.0~\cite{sdxl} as our generator, and prompt it with ``a photo of a single and fully visible [class name]''. The emphasis on ``single and fully visible'' is crucial, as the model might otherwise generate images with multiple objects or encounter occlusion issues, as observed in~\cite{x-paste}. The [class name] represents a specific category from 80 classes of COCO~\cite{coco}.

\subsubsection{Isolating Objects from the Background}
For ease in subsequent processes, we need to separate the objects from the backgrounds where they are embedded. To accomplish this, we first utilize an open-set detector, Grounding-DINO~\cite{grounding-dino} to outline the regions containing the objects. Subsequently, we prompt SAM~\cite{sam} with this bounding box as to obtain the final segmentation results. Thus far, we have established a library containing instances from various categories. These instances are background-free, allowing for composition on a blank canvas, while their attributes and relations can be controlled.

\subsubsection{Arranging Objects on the Canvas}
Recall that our goal is to manipulate a specific visual-linguistic concept of an image. Currently, this task appears straightforward when we exclude the background from consideration. We can retrieve instances from the library and arrange them on a blank canvas according to our specifications. For example, we can flexibly control the quantity of an object through duplication operations, modify their sizes via resizing, and determine whether an object exists and specify the position of an existing object. This process resembles the concept of copy-paste~\cite{copy-paste, x-paste}, with a notable distinction: we paste objects onto a blank background and placing emphasis on controlling properties such as the size, position, existence, and quantity of each individual object.

\subsubsection{Infilling the Missing Background}
\label{sec:bg_fill_main}
\begin{figure}
    \centering
    \includegraphics[width=0.95\linewidth]{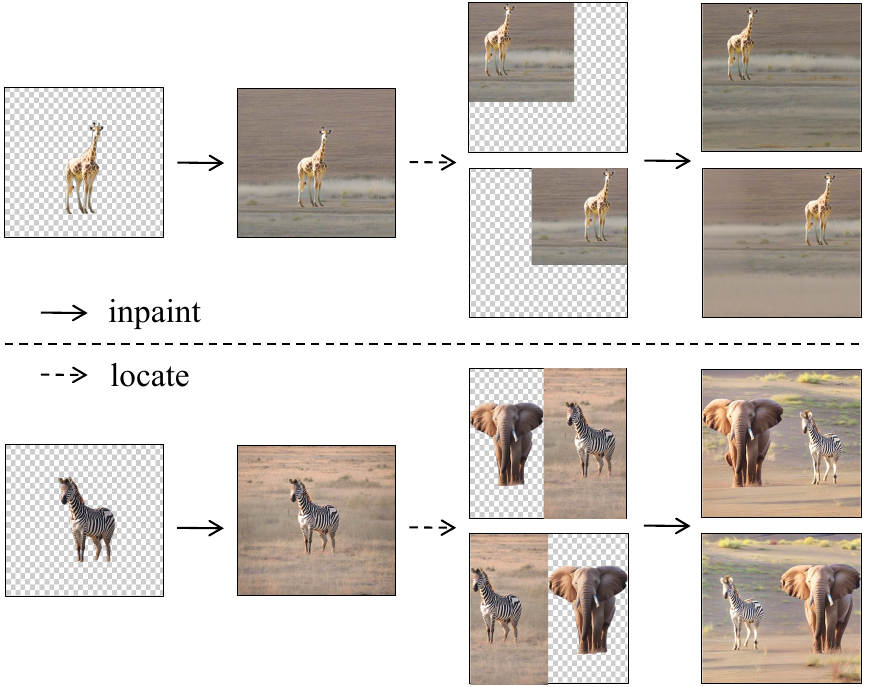}
    \caption{\textbf{Consistent background inpainting strategy.} We first generate an initial background shared by all candidate images. Then, we expand around this region, ensuring consistency in the backgrounds of different images.}
    \label{fig:background_fill}
\end{figure}

We have constructed images with differences in specified attributes, however, they currently lack a suitable background. To fill the missing area, we employ a inpainting model~\cite{sdxl} which demonstrates proficiency in filling large holes. It is worth noting that generating backgrounds individually for each image would result in significant differences in the backgrounds, posing a challenge to maintaining consistency among candidate images. To overcome this, we introduce a strategy where images within the same candidate set share a common and consistent background during the inpainting process.
As depicted in the upper part of~\cref{fig:background_fill}, to present the giraffe in different positions, we start by surrounding it with an initial background. Then, we relocate this initial image on the canvas as required, and fill the remaining blank space. Similarly, in the lower part, we first embed the zebra into a reasonable environment. Following that, we expand the scene horizontally, introduce the elephant, and fill the blank areas through inpainting. In summary, we begin the process by generating an initial background, which is then expanded to the surroundings, effectively ensuring consistency among the candidate images. Additional examples, such as adjusting the size or quantity of an object, can be found in the supplementary.

\subsection{SPEC Benchmark}
\label{sec:spec_benchmark}
Utilizing the data engine outlined in~\cref{sec:data_pipeline}, we carefully devise the SPEC benchmark with the goal of assessing the  performance of VLMs in comprehending object size, position, existence and count. An overview of the SPEC benchmark is presented in \cref{fig:spec_overview}. SPEC contains six subsets, which will be elaborated as follows:

\begin{figure*}
    \centering
    \includegraphics[width=\linewidth]{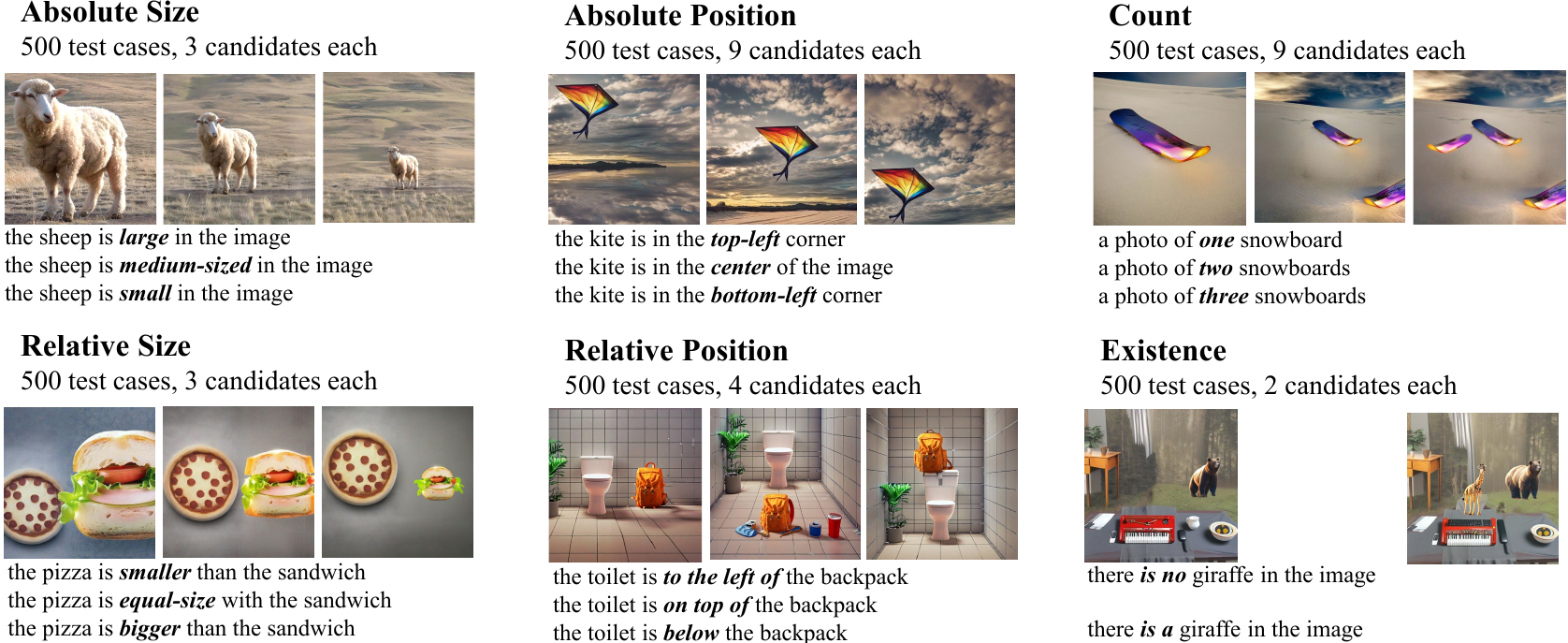}
    \caption{\textbf{An overview of the SPEC benchmark.} SPEC consists of six distinct subsets, distributed across the dimensions of Size, Position, Existence and Count. Each test case consists of an image candidate set, which differs only in certain visual concept, and a text candidate set, which differs only in corresponding language concept. Due to space constraints, we present a maximum of three images and texts here, however, more comprehensive test cases are available in the supplementary material.}
    \label{fig:spec_overview}
\end{figure*}

\fakeparagraph{Absolute Size}
reflects how large an object is in comparison to the entire image. We categorize this into three level: large, medium, or small, and define them following:
\begin{equation}
    \text{Size}_{abs.}(x) =
    \begin{cases}
        \texttt{small},  & {P} \leq 0.2 \\
        \texttt{medium}, & 0.4 \leq {P} \leq 0.6 \\
        \texttt{large},  & {P} \geq 0.8,\\
    \end{cases}
    \label{eq:absolute_size}
\end{equation}
where $P$ denotes the proportion of the space occupied by the object relative to the area of the entire image. A safety threshold is deliberately introduced between these three levels to prevent ambiguity.

\fakeparagraph{Relative Size} focuses on the size relationship between two objects. It is categorized as follows: object A is [smaller than, equal to, larger than] object B, and we measure this following:
\begin{equation}
    \text{Size}_{rea.}(A, B) =
    \begin{cases}
        A~\texttt{smaller~than}~B,  & {R} \leq 0.5 \\
        A~\texttt{equal~to}~B, & 0.9 \leq {R} \leq 1.1 \\
        A~\texttt{larger~than}~B,  & {R} \geq 2,
    \end{cases}
    \label{eq:relative_size}
\end{equation}
where $R=\frac{S_A}{S_B}$ is the ratio of the areas of object A and object B.

\fakeparagraph{Absolute Position} signifies the location of an object relative to the image. We partition the image into a 3$\times$3 grid, defining nine possible positions: top-left, top, top-right, left, center, right, bottom-left, bottom, and bottom-right. The absolute position of an object is determined based on the grid in which its center point resides.

\fakeparagraph{Relative Position} describes the spatial relationship between two objects. We consider four common spatial relationships: A is [to the left of, to the right of, above, below] B. The position relationship between objects is defined based on the relative positions of their center points.

\fakeparagraph{Existence} indicates whether an object appears in a given image, expressed using existential quantifiers: There is [no, at least one] object in the image.

\fakeparagraph{Count} represents the number of occurrences of an object, providing a metric for the model's quantitative understanding.  Due to potential occlusion issues with a large number of objects, we restrict our consideration to the range of 1 to 9: ``there are [one, two, $\cdots$, nine] object(s) in the image''.

\fakeparagraph{Data Format.}
The basic unit of SPEC is an individual test case, wherein each test case comprises two components: an image candidate set, which differs only in certain visual aspects, and a text candidate set, which differs only in the corresponding language descriptions. We formally represent a test case as:
\begin{equation}
    \mathbb{T}: \left(\mathcal{I}=\{I_1, \cdots, I_K\}, \mathcal{T}=\{T_1, \cdots, T_K\}\right),
    \label{eq:test_case}
\end{equation}
where $\mathcal{I}$ and $\mathcal{T}$ represent the image and text candidate sets, respectively. The i-th image $I_i$ is paired with the i-th text $T_i$, \ie, they mutually describe each other. $K$ is the semantic cardinality of the test case, determined by the definition of each subset. For instance, if a test case belongs to the absolute size subset, then $K=3$~(representing the three semantics: large, medium, small). In~\cref{fig:spec_overview}, we present examplar test cases for each subset, and more examples can be found in the supplementary.

\fakeparagraph{Comparing SPEC with other benchmarks.}
\begin{table}
\centering
\resizebox{\columnwidth}{!}{%
\renewcommand\arraystretch{1.2}
\begin{tabular}{lccccc}
\toprule
Benchmarks                               & Visual HN  & Scalability &Text-realistic& Photo-realistic & \#Candidates \\ \cmidrule(lr){1-6}
VALSE~\cite{VALSE}                       & \XSolidBrush & \Checkmark  & \Checkmark & \Checkmark &  2    \\
ARO~\cite{bow_aro}                       & \XSolidBrush & \Checkmark  &\XSolidBrush& \Checkmark &  2    \\
Winoground~\cite{thrush2022winoground}   & \Checkmark   & \XSolidBrush& \Checkmark & \Checkmark &  2    \\
Eqben~\cite{Eqben}                       & \Checkmark   & \Checkmark  & \Checkmark & \HalfCheckmark &  2    \\
\rowcolor[RGB]{230, 230, 230}
SPEC(Ours)                               & \Checkmark   & \Checkmark  & \Checkmark & \Checkmark &  2-9  \\
\bottomrule
\end{tabular}%
}
\caption{Comparison of SPEC with other fine-grained benchmarks.}
\label{tab:compare_spec}
\end{table}
In~\cref{tab:compare_spec}, we conduct a comprehensive comparison of SPEC with four similar benchmarks. \textbf{Visual HN} indicates the presence of image hard negatives, which are essential for the text-to-image matching task. Notably, VALSE~\cite{VALSE} and ARO~\cite{bow_aro} concentrate solely on text hard negatives, neglecting their visual counterparts. \textbf{Scalability} assesses whether the data construction method can be scaled up. For instance, Winoground~\cite{thrush2022winoground} is limited to 400 examples due to high costs for manual collection. Additionally, \textbf{Text-realistic} evaluates the grammatical correctness of the texts, where ARO involves directly swapping the positions of two words without considering grammar. Similarly, \textbf{Image-realistic} indicates the realism of the images. Eqben~\cite{Eqben} incorporates images rendered using a virtual engine, compromising their quality. In contrast, images from SPEC, while synthesized, leverage advanced image generation models and our effective data construction pipeline, resulting in photo-realistic images. Finally, we compare the number of \textbf{Candidates} in each test case. In all datasets except SPEC, each example features only two candidates, \ie, identifying the correct item from two candidates, which is relatively straightforward. In contrast, tasks in SPEC are more challenging, and the semantic space covered by the candidate set is more comprehensive. For example, in the case of relative position, the candidates include all four semantic directions (up, down, left, right), and in absolute position and count, it extends to nine candidates. As will be discussed below, more confusing candidates can be readily used as hard negatives to improve current VLMs. 

\begin{table*}[]
\centering
\resizebox{\linewidth}{!}{%
\renewcommand\arraystretch{1.2}
\begin{tabular}{lccccccccccccccccccc}
\toprule
& \multicolumn{3}{c}{Absolute Size} & \multicolumn{3}{c}{Relative Size} & \multicolumn{3}{c}{Absolute Position} & \multicolumn{3}{c}{Relative Position} & \multicolumn{3}{c}{Existence} & \multicolumn{3}{c}{Count} \\ 
\cmidrule(lr){2-4} \cmidrule(lr){5-7} \cmidrule(lr){8-10} \cmidrule(lr){11-13} \cmidrule(lr){14-16} \cmidrule(lr){17-19}   
& \textsc{i}2\textsc{t} & \textsc{t}2\textsc{i} & \textsc{cls} & \textsc{i}2\textsc{t} & \textsc{t}2\textsc{i} & \textsc{cls} & \textsc{i}2\textsc{t} & \textsc{t}2\textsc{i} & \textsc{cls} &  \textsc{i}2\textsc{t} & \textsc{t}2\textsc{i} & \textsc{cls} & \textsc{i}2\textsc{t}  & \textsc{t}2\textsc{i} &  \textsc{cls} & \textsc{i}2\textsc{t} & \textsc{t}2\textsc{i} & \textsc{cls} \\ \midrule
Random                       &  33.3 & 33.3 & 1.3 &  33.3 & 33.3 & 2.5 & 11.1  & 11.1 & 1.3 & 25.0  & 25.0 & 2.5  & 50.0 & 50.0 & 5.0  & 11.1  & 11.1 & 1.3 \\
FlAVA~\cite{singh2022flava}  &  37.3 & 37.2 & 89.3&  32.9 & 32.3 & 84.1& 13.1  & 15.7 & 88.8& 25.5  & 26.7 & 84.0 & 57.9 & 51.9 & 74.4 & 14.4  & 21.2  & 75.8  \\
BLIP~\cite{li2022blip}       &  43.3 & 42.7 & 97.7&  33.2 & 32.5 & 98.2& 12.2  & 11.0 & 97.9& 30.5  & 29.7 & 97.6 & 55.4 & 50.1 & 96.0 & 37.4  & 37.1  & 93.8  \\
CoCa~\cite{CoCa}             & 39.1  & 36.1 & 95.9&  33.6 & 33.3 & 95.4& 11.7  & 11.8 & 96.1& 30.3  & 28.8 & 92.2 & 50.9 & 50.0 & 83.7 & 36.5  & 35.6  & 89.6  \\
CLIP~\cite{clip}             & 42.5  & 36.1 & 92.1&  34.4 & 33.4 & 94.9& 12.6  & 12.3 & 92.8& 28.0  & 26.6 & 90.1 & 58.3 & 51.2 & 84.0 & 25.1  & 23.2  & 83.9  \\
\rowcolor[RGB]{230, 230, 230}
Ours                         & 68.9  & 60.7 & 96.3&  40.3 & 44.1 & 97.3& 30.6  & 34.2 & 96.9& 46.6  & 46.9 & 96.2 & 83.4 & 53.1 & 92.5 & 55.6 & 57.8  & 92.5  \\
\bottomrule
\end{tabular}%
}
\smallskip
\caption{\textbf{Evaluation results on SPEC.} We extensively benchmark four state-of-the-art VLMs on SPEC to investigate their comprehension of fine-grained visual-linguistic concepts. Our evaluation employ two metrics, $\textsc{I2Tacc}$ and $\textsc{T2Iacc}$, and we report the detailed performance on the each subset. We also report the classification accuracy, $\textsc{cls}$, to highlight the capability in recognizing object categories. The first row indicates accuracy at chance level, serving as a baseline for comparison. }
\label{tab:main_spec_result}
\end{table*}

\section{Diagnose: Probing VLMs on SPEC}
\label{sec:expriment}
We conduct a systematical evaluation of four state-of-the-art VLMs: CLIP~\cite{clip}, BLIP~\cite{li2022blip}, FLAVA~\cite{singh2022flava} and CoCa~\cite{CoCa} using our newly proposed SPEC benchmark, aiming to uncover that to what extent and in what aspect are they excelling or suffering.

\subsection{Evaluation Task and Protocol}
\fakeparagraph{Symmetric image-text matching task.} We conduct evaluations using the image-text matching task, where each test case $\mathbb{T}\text{=}(\mathcal{I}, \mathcal{T})$ comprises an image set $\mathcal{I}$ and a corresponding text set $\mathcal{T}$, as outlined in~\cref{eq:test_case}. The evaluation process is symmetric, considering both visual and textual perspectives. In the image-to-text matching task, when querying with an image $I_i$, the model is required to accurately identify $T_i$ from the candidate set $\mathcal{T}$. Similarly, in the text-to-image matching task, the goal is to find $I_i$ for a given query text $T_i$. In practice, we accomplish the matching process using the similarity $s(I, T)$ between a given image $I$ and text $T$. A candidate will be selected if its similarity with the query ranks first among the entire candidate set.

\fakeparagraph{Evaluation protocols.} We measure the performance on SPEC using two metrics: $\textsc{I}2\textsc{Tacc}$ and $\textsc{T}2\textsc{Iacc}$, representing the accuracy of image-to-text and text-to-image matching task, respectively:

\begin{equation}
    \textsc{I}2\textsc{Tacc}=\frac{1}{|D|}\sum\limits_{(\mathcal{I}_i, \mathcal{T}_i) \in D} \frac{1}{|\mathcal{I}_i|} \sum\limits_{I_j \in \mathcal{I}_i} h(I_j, \mathcal{T}_i)
    \label{eq:i2t_acc}
\end{equation}

\begin{equation}
    \textsc{T}2\textsc{Iacc}=\frac{1}{|D|}\sum\limits_{(\mathcal{I}_i, \mathcal{T}_i) \in D} \frac{1}{|\mathcal{T}_i|} \sum\limits_{T_j \in \mathcal{T}_i} g(T_j, \mathcal{I}_i),
    \label{eq:t2i_acc}
\end{equation}
where $D$ contains $|D|$ test cases, $h(I_j, \mathcal{T}_i)$ equals to 1 if and only if $I_j$ correctly find its matched text $T_j$ from the candidate set $\mathcal{T}_i$, otherwise it is set to 0. Similarly,  $g(T_j, \mathcal{I}_i)$ equals 1 if and only if $T_j$ correctly finds its matched image $I_j$ from the candidate set $\mathcal{I}_i$, otherwise, it is set to 0.

\subsection{Key Insights from SPEC Results}
We evaluate four VLMs using the SPEC benchmark, and their results are summarized in~\cref{tab:main_spec_result}. We find that all the models exhibit a limited accuracy close to random chance, from which we gain the following insights:

\fakeparagraph{Even state-of-the-art VLMs perform at chance level.}
From the results, we surprisingly find that even the most advanced VLMs achieve only a marginal advantage compared to random chance, which sharply contrasts with their impressive performance on common tasks. For instance, CLIP~\cite{clip} demonstrates a mere 33.4$\%$ $\textsc{t2iacc}$ for relative size recognition, while the chance-level accuracy is 33.3$\%$.
While BLIP performs reasonably well in absolute size, surpassing random level by around 9.7$\%$, the $\textsc{i2tacc}$ on relative size is 0.7$\%$ lagged behind. CoCa~\cite{CoCa} and FLAVA~\cite{singh2022flava} also exhibit significant weaknesses in performance.
The last row presents the performance of our improved model, demonstrating significant advancements across all metrics (as will be introduced in~\cref{sec:remedy}).

\fakeparagraph{The challenge arises from the task itself, not the data.}
One might attribute the poor performance of VLMs to the data quality or distribution. To address this concern, we conduct an additional experiment.
Specifically, we perform classification experiments using the SPEC dataset to assess the model's understanding of nouns or object categories. In this context, the models exhibit impressive performance, achieving approximately 90$\%$ Top-1 accuracy in the 80-class classification task~(denoted as $\textsc{cls}$ in~\cref{tab:main_spec_result}). This aligns well with earlier findings~\cite{bow_aro} that VLMs struggle in compositional reasoning while excelling in object category recognition. 
The remarkable accuracy of the models in the object classification task confirms the high quality of SPEC data. This also validates that the challenges faced by VLMs stem from the tasks which require fine-grained recognition rather than issues with the data itself. 

\subsection{Discussion on Model Limitations}
We attribute the poor performance of vision and language models on SPEC to to their pretraining methods, specifically, the inherent limitation in standard contrastive loss. The conventional contrastive learning involves randomly sampling batches of images and texts, requiring the model to identify matching pairs within the batch. This task is intended to facilitate alignment between text and image spaces. However, as highlighted in prior studies~\cite{bow_aro, beyoundNouns}, the substantial differences between items in a randomly sampled batch allows the model to effortlessly complete this task. It can easily achieves this by focusing solely on nouns in the text and object categories in the images through a shortcut~\cite{shortcut}. This leads the model biased towards noun concepts, neglecting other finer-grained concepts. This is the reason why these models demonstrate poor performance on fine-grained tasks that demanding understanding concepts beyond nouns.

\section{Optimize: A Simple but Effective Remedy}
\label{sec:remedy}
We experiment with CLIP~\cite{clip} and propose a remedy to enhance its performance in fine-grained understanding.

\subsection{Method}
CLIP~\cite{clip} consists of an visual encoder to extract image embedding: $e_I=\mathcal{E}_I(I)$ and a textual encoder to extract text embedding: $e_T=\mathcal{E}_T(T)$. The similarity score between an image $I$ and a text $T$ is computed following:
\begin{equation}
    s(I, T)=\text{exp}\left(\frac{\tau e_I^Te_T}{\|e_I\|^2\|e_T\|^2}\right),
    \label{eq:cos_sim}
\end{equation}
where $\tau$ is a learnable temperature.

In order to guide CLIP to focus on fine-grained visual-linguistic concepts, we incorporate confusing images and text as hard negatives within the same batch. This requires CLIP to pull positives closer and push hard negatives away, thereby enhancing its ability to discern nuanced visual and textual differences. Specifically, we introduce an hard negative aware contrastive loss $\mathcal{L}_{hn}=\mathcal{L}_{hn}^{\textsc{i2t}}+\mathcal{L}_{hn}^{\textsc{t2i}}$, which comprises an image-to-text and a text-to-image term:
\begin{equation}
    \mathcal{L}_{hn}^{\textsc{i2t}}=-\sum\limits_{i}\text{log}\frac{s(I_i,T_i)}{\sum\limits_{T_j\in\mathcal{T}}s(I_i,T_j)+\sum\limits_{T_k^{hn}\in\mathcal{T}^{hn}}s(I_i,T_k^{hn})}
    \label{eq:loss_hn_i2t}
\end{equation}

\begin{equation}
    \mathcal{L}_{hn}^{\textsc{t2i}}=-\sum\limits_{i}\text{log}\frac{s(I_i,T_i)}{\sum\limits_{I_j\in\mathcal{I}}s(I_j,T_i)+\sum\limits_{I_k^{hn}\in\mathcal{I}^{hn}}s(I_k^{hn},T_i)},
    \label{eq:loss_hn_t2i}
\end{equation}
where $\mathcal{I}$ and $\mathcal{T}$ represent the trivial images and texts, respectively, while $\mathcal{I}^{hn}$ and $\mathcal{T}^{hn}$ denote non-trivial hard negatives. In our implementation, these hard negative examples are constructed using the data pipeline described in~\cref{sec:data_pipeline}, and more details are in the supplementary.

To preserve the inherent zero-shot capability of CLIP, we also leverage the conventional image-text pairs from LAION-400M~\cite{laion400m}. We apply the standard contrastive loss of CLIP~\cite{clip} to these data, introducing an additional loss term $\mathcal{L}_{clip}$. The overall loss consists of two terms:
\begin{equation}
    \mathcal{L} = \mathcal{L}_{clip} + \lambda\mathcal{L}_{hn},
    \label{eq:loss_total}
\end{equation}
where $\lambda$ is a hyperparameter that balancing these two terms.
Training on this multi-task loss enables improving the performance of CLIP in fine-grained understanding while maintaining its zero-shot capability.

\subsection{Experiments}
\fakeparagraph{Training details.} We experiment with the ViT-B/32 variant of CLIP, and resume from the OpenAI pretrained checkpoint~\cite{clip}. We finetune for 1,000 steps using a cosine schedule with an initial learning rate of $1e\text{-}6$ and use 800 steps for warm up. The batch size of LAION data is set to 2048, and the batch size of hard negative data is set to 768. The weight $\lambda$ of the hard negative aware loss is set to 0.2.

\fakeparagraph{Main results.} 
We utilize the SPEC benchmark to assess the understanding of model in fine-grained concepts. In \cref{tab:main_result}, we present the average $\textsc{I2Tacc}$ and $\textsc{T2Iacc}$ on all subsets of SPEC.
To assess the general performance of the model, we also utilize the toolkit from ELEVATER~\cite{Li2022ELEVATERAB} to evaluate the zero-shot performance on 9 classification and retrieval datasets, and report the average accuracy. Compared to CLIP~\cite{clip} , our model demonstrates remarkable advancements with a 19.8$\%$ boost in $\textsc{i2tacc}$, an 18.9$\%$ improvement in $\textsc{t2iacc}$ on SPEC, and a noteworthy 1.2$\%$ enhancement in zero-shot accuracy.
We also conduct ablation on different training configurations. From the results in~\cref{tab:main_result}, it can be observed that the introducing of $\mathcal{L}_{hn}$ significantly improves the performance on SPEC. Moreover, the $\mathcal{L}_{clip}$ plays a crucial role in preserving zero-shot performance. Without it, we observe a decline in accuracy by 5.1$\%$. With the combination of these two losses, we achieve substantial improvement in SPEC while maintaining the original zero-shot capability.

\begin{table}
\centering
\resizebox{0.95\columnwidth}{!}{%
\renewcommand\arraystretch{1.2}
\begin{tabular}{lccccc}
\toprule
&\multicolumn{2}{c}{Config}&\multicolumn{2}{c}{SPEC}&Zero-shot \\
\cmidrule(lr){2-3} \cmidrule(lr){4-5} \cmidrule(lr){6-6}
& $\mathcal{L}_{clip}$ & $\mathcal{L}_{hn}$ &  \textsc{i}2\textsc{t} &  \textsc{t}2\textsc{i}  & Accuracy \\ \cmidrule(lr){2-6}
CLIP                                             &             &             & 33.5  & 30.5  &  67.5     \\
+ $\mathcal{L}_{hn}$                             &             &  \checkmark & 64.3  & 60.8  &  62.4     \\
+ $\mathcal{L}_{clip}$                           & \checkmark  &             & 32.2  & 31.5  &  69.4     \\
\rowcolor[RGB]{230, 230, 230}
+ $\mathcal{L}_{hn}$+$\mathcal{L}_{clip}$~(ours) & \checkmark  &  \checkmark & 53.3  & 49.4  &  68.7     \\
\bottomrule
\end{tabular}%
}
\caption{\textbf{Main Results:} The first row represents the pretrained checkpoint without fine-tuning. We sequentially introduce two additional loss terms to investigate their impact for the performance.}
\label{tab:main_result}
\end{table}

\fakeparagraph{Validation on other fine-grained benchmarks.} 
To assess whether our approach aids the model in acquiring fundamental visual-linguistic understanding or merely leads to overfitting on SPEC, we conduct evaluations on two additional benchmarks which also focus on the assessment of fine-grained concepts. ARO~\cite{bow_aro} explores three aspects of vision-language understanding: object attributes, inter-object relations, and word ordering. Eqben~\cite{Eqben} focuses on minimal visual semantic changes, aiming to diagnose VLMs in understanding fine-grained concepts such as counting and location. In \cref{tab:cross_validate}, we present the experimental results, demonstrating a clear improvement compared to CLIP~\cite{clip} on both datasets, For example, compared to CLIP, our method shows an average improvement of 2$\%$ on Eqben and respective enhancements of 3.2$\%$, 9.8$\%$, and 7.4$\%$ on the three subsets of ARO. The consistent improvement on these datasets demonstrates that our approach has facilitated the model in acquiring transferable fine-grained understanding capabilities.

\begin{table}
\centering
\resizebox{\columnwidth}{!}{%
\renewcommand\arraystretch{1.2}
\begin{tabular}{lcccccc}
\toprule
&\multicolumn{3}{c}{Eqben}&\multicolumn{3}{c}{ARO}\\
\cmidrule(lr){2-4} \cmidrule(lr){5-7}
                   & Image & Text  & Group & Attribute & Relation & Order \\ \cmidrule(lr){2-7}
CLIP               & 17.6  & 21.4  & 10.1  & 63.2      & 63.9     & 53.3  \\
$\text{CLIP}_{FT}$ & 18.1  & 23.7  & 11.1  & 65.1      & 68.0     & 54.1  \\
\rowcolor[RGB]{230, 230, 230}
Ours               & 19.5  & 24.0  & 11.7  & 66.4      & 73.7     & 60.7  \\
\bottomrule
\end{tabular}%
}
\caption{\textbf{Cross-dataset evaluation results on Eqben~\cite{Eqben} and ARO~\cite{bow_aro}.} To demonstrate that the improvement comes from the negative loss, rather than training on more data, we also report $\text{CLIP}_{FT}$, which is also finetuned but without negative samples.}
\label{tab:cross_validate}
\end{table}

\section{Conclusion}
\label{sec:conclusion}
In this study, we explored the comprehension abilities of Visual Language Models (VLMs) with respect to fine-grained visual-linguistic concepts. We first established an efficient pipeline to synthesize candidate images that exclusively differ in a particular visual attribute. Leveraging this pipeline, we created the SPEC benchmark to diagnose the comprehension proficiency of VLMs in terms of object size, position, existence, and count. Upon evaluating four leading VLMs using SPEC, we uncovered substantial performance limitations. To address this, we introduced an enhancement strategy that effectively optimizes the model for fine-grained understanding, while maintaining its original zero-shot capability.

\noindent\textbf{Acknowledgement} This project was supported by NSFC under Grant No. 62102092.

{
    \small
    \bibliographystyle{ieeenat_fullname}
    \bibliography{main}
}
\end{document}